\documentclass{article}

\usepackage[preprint]{neurips_2026}

\usepackage[utf8]{inputenc}
\usepackage[T1]{fontenc}
\usepackage{hyperref}
\usepackage{url}
\usepackage{booktabs}
\usepackage{amsfonts}
\usepackage{nicefrac}
\usepackage{microtype}
\usepackage{xcolor}
\newcommand{\anonrepo}{\href{https://github.com/YahyaAalaila/HawkesNest}{\texttt{HawkesNest repository}}}
\usepackage[small]{caption}
\usepackage{graphicx}
\usepackage{amsmath}
\usepackage{amsthm}
\usepackage{algorithm}
\usepackage{algorithmic}
\usepackage{subcaption}
\usepackage{makecell}
\usepackage{multirow}
\usepackage{amssymb}
\usepackage{natbib}
\usepackage{tikz}
\usetikzlibrary{arrows.meta}
\usepackage[most]{tcolorbox}
\usepackage{listings}
\usepackage{xcolor}

\definecolor{codebg}{RGB}{245,245,245}
\definecolor{codeframe}{RGB}{220,220,220}

\newtcblisting{cmdblock}{
  colback=codebg,
  colframe=codeframe,
  boxrule=0.3pt,
  arc=2pt,
  left=6pt,
  right=6pt,
  top=6pt,
  bottom=6pt,
  listing only,
  breakable,
  enhanced
}

\title{HawkesNest:\\ A Multi-Axis Synthetic Benchmark for Spatiotemporal Pattern Complexity}

\author{%
  Yahya Aalaila$^{1, 2}$ \quad
  Sumantrak Mukherjee$^{1}$ \quad
  Gerrit Gro\ss{}mann$^{1}$ \quad
  Sebastian Vollmer$^{1,2}$ \\[0.6em]
  $^{1}$German Research Center for Artificial Intelligence (DFKI), \\
  \hspace*{1.2em}Data Science and its Applications Research Group, Kaiserslautern, Germany \\
  $^{2}$Department of Computer Science, Rhineland-Palatinate Technical University \\
  \hspace*{1.2em}of Kaiserslautern-Landau (RPTU), Kaiserslautern, Germany \\[0.4em]
  \texttt{yahya.aalaila@dfki.de} \quad
  \texttt{sumantrak.mukherjee@dfki.de} \\
  \texttt{gerrit.grossmann@dfki.de} \quad
  \texttt{sebastian.vollmer@dfki.de}
}

\begin{document}

\maketitle
\begin{abstract}
Evaluation of spatiotemporal point process (STPP) models relies heavily on opaque real-world datasets, where latent generative structure is unknown and model failures are difficult to attribute. We introduce HawkesNest, a generator-aligned benchmark for controlled spatiotemporal pattern complexity built on a multivariate Hawkes backbone. 
HawkesNest defines four complexity axes: space--time entanglement, background heterogeneity, cross-type interaction, and domain topology. Each axis is associated with a deterministic index computed from the latent data-generating mechanism.
By varying these axes while holding global rate, stability, and simulation budget fixed, HawkesNest enables diagnostic stress tests of STPP models under known structural difficulty. We verify that the indices are monotone and nearly orthogonal under controlled sweeps. We illustrate its use by showing that Hawkes-family baselines degrade under joint heterogeneity--entanglement complexity, even though they are structurally aligned with the Hawkes data-generating backbone. We further show that HawkesNest exposes neural-model sensitivity: AutoSTPP remains vulnerable under isolated increases in space--time entanglement.
\paragraph{Code.} Available at \anonrepo.
\end{abstract}
\section{Introduction}
Real-world phenomena such as crime, epidemics, earthquakes, and traffic incidents can be represented as spatiotemporal point patterns. Accordingly, spatiotemporal point process (STPP) models have seen rapid development~\citep{moller1998log, miscouridou2022cox,semenova2022priorvae, mukherjee2025neural}. Yet rigorous evaluation remains difficult because most benchmarks are built on opaque real-world corpora whose latent generative structure is unknown. In such datasets, multiple spatiotemporal mechanisms may coexist, making it difficult to determine whether improvements in likelihood or forecasting accuracy reflect genuine pattern recovery or sensitivity to dataset-specific artifacts~\citep{zhang2025strapspatiotemporalpatternretrieval,han2025retrievalaugmentedtimeseries,atluri2018spatio}.

Beyond standard temporal patterns such as trend and seasonality, spatiotemporal data exhibit structure tied to the joint evolution of space and time. One example is space--time entanglement: a parent event may produce nearby offspring shortly after it occurs, but more dispersed offspring after a longer parent--child time lag, so the spatial triggering pattern is coupled to time, violating the separability assumptions often used in STPP models~\citep{li2025ssl}. Background heterogeneity arises when persistent spatial or temporal variation changes the baseline event rate before any self-excitation is considered; for example, a commercial zone may dominate during working hours while residential areas dominate overnight~\citep{li2025ssl}.

Many applications also impose domain constraints: traffic accidents occur on road networks, sewer overflows propagate along river basins, and earthquakes follow fault systems~\citep{topo}. In such settings, events evolve on lower-dimensional substrates and interact through restrictive, non-Euclidean geometry~\citep{regazzoni2024learning}.

Existing synthetic generators partially address this gap, but typically vary a single mechanism or operate in low-dimensional settings, limiting their diagnostic value under combined spatiotemporal complexity. \textsc{EasyTPP}~\citep{xue2024easytpp}, for example, provides a univariate Hawkes sequence without spatial structure or event marks. The pinwheel toy used in neural STPP tutorials~\citep{chen2021neuralstpp} contains a visible spatial pattern, but does not provide systematic controls for temporal heterogeneity, cross-type interaction, or domain constraints. ETAS simulators~\citep{ogata1998space,jalilian2019package} are tailored to earthquake-specific triggering assumptions, and crime hot-spot simulations~\citep{mohler2011self} vary baseline intensity while keeping contagion structure fixed.

\paragraph{Contributions.}
These limitations motivate a diagnostic benchmark that can jointly and systematically control multiple sources of spatiotemporal complexity under one generative process. Our contributions are:

\begin{itemize}
    \item \textbf{Generator-aligned complexity axes.} We formalize four pillars of spatiotemporal complexity: space--time entanglement, background heterogeneity, cross-type interactions, and domain topology. We associate each with a deterministic index satisfying baseline minimality and monotonic control.

    \item \textbf{A dialable Hawkes-based simulator.} We introduce a modular multivariate Hawkes backbone whose background, triggering, interaction, and domain components can be composed while preserving global rate and stability constraints.

    \item \textbf{Diagnostic validation across model families.} We verify monotonicity and near-orthogonality of the indices under controlled sweeps. We then show that Hawkes-family baselines degrade under joint heterogeneity--entanglement complexity despite their structural alignment with the generator, and that AutoSTPP remains sensitive to isolated space--time entanglement.
\end{itemize}

\section{Background}\label{section:back}

A space--time Hawkes process models event occurrences in a spatial domain
$\mathcal S\subseteq\mathbb R^d$ over a finite time horizon~\citep{miscouridou2022cox,mukherjee2025neural}. Let
$\mathcal H_t=\{(\mathbf{s}_i,t_i):t_i<t\}$ denote the history of events strictly before time $t$.

\paragraph{Conditional intensity.}
The process is specified by its conditional intensity
$\lambda(\mathbf{s},t\mid\mathcal H_t)$, the infinitesimal expected rate of an
event at location $\mathbf{s}$ and time $t$ given the past. For readability, we
omit $\mathcal H_t$ below:
\begin{equation}\label{eq:hawkes-intensity}
\begin{split}
\lambda(\mathbf{s},t)
=\;
\underbrace{\mu(\mathbf{s},t)}_{\text{background}}
+
\sum_{j:\,t_j<t}
\underbrace{\phi\!\,\bigl(\mathbf{s}-\mathbf{s}_j,\;t-t_j\bigr)}_{\text{triggering kernel}} .
\end{split}
\end{equation}
Here, $\mu:\mathcal S\times[0,T]\to\mathbb R_{\ge 0}$ is the background rate,
capturing exogenous heterogeneity in space and time, and
$\phi:\mathcal S\times(0,\infty)\to\mathbb R_{\ge 0}$ is the triggering kernel,
quantifying how a past event at $\mathbf{s}_j$ and lag $t-t_j$ increases the
risk of a new event at location $\mathbf{s}$.

\paragraph{Stability.}
The offspring mean
\[
\eta
=
\int_{0}^{\infty}\!\!\int_{\mathcal S}
\phi(\mathbf s,\tau)\,d\mathbf s\,d\tau
\]
is the expected number of offspring generated by one event. For a single-type process, stability requires $\eta<1$. In the multitype case, the branching matrix \(A\) must satisfy $\rho(A) < 1$,
where \(\rho(A)\) denotes the spectral radius. In our simulator, we enforce a conservative bound through global rescaling~\citep{baddeley2016spatial,daley2003introduction}.

\paragraph{Background intensity.}\hypertarget{par:bg}{}
The background term $\mu(\mathbf{s},t)$ is the exogenous arrival rate, independent
of self-excitation. Common choices include: (i) a constant baseline
$\mu(\mathbf{s},t)\equiv c$, yielding a stationary homogeneous Poisson ground
process; (ii) a spatially varying baseline $\mu(\mathbf{s},t)=c f_s(\mathbf{s})$,
where $f_s$ is a bounded non-negative surface such as a Gaussian mixture; and
(iii) a spatiotemporal baseline $\mu(\mathbf{s},t)=c f_s(\mathbf{s})f_t(t)$, or
a fully joint field $\mu(\mathbf{s},t)=c f_{st}(\mathbf{s},t)$, capturing cycles,
seasonality, or regime shifts.

\paragraph{Triggering kernel.}\hypertarget{par:trig_kernel}{}
The kernel encodes how a past event influences future risk as a function of
spatial lag and temporal lag $\tau>0$. A separable kernel
$\phi(\mathbf{s},\tau)=g(\tau)h(\lVert\mathbf{s}\rVert)$, with exponential $g$
and Gaussian $h$, recovers the prototype structure used in ETAS-style earthquake
models and many crime Hawkes models. More general entangled kernels allow
$\phi$ to depend jointly on space and time, e.g.
$\phi(\mathbf{s},\tau)=\kappa(\lVert\mathbf{s}\rVert,\tau)$, capturing cases
where spatial dispersion and temporal decay are coupled~\citep{helmstetter2003foreshocks}.

\paragraph{Multi-mark extension.}\hypertarget{par:multi_mark}{}
Real datasets often contain event marks or labels. In the marked case, the
single-event intensity in Equation~\eqref{eq:hawkes-intensity} becomes one
component per mark, $\lambda_m(\mathbf s,t)$, each with its own background
$\mu_m(\mathbf s,t)$. Every past event of type $n$ contributes to each
$\lambda_m$ through a cross-triggering kernel $\phi_{mn}(\mathbf s,t)$:
diagonal kernels $(m=n)$ capture self-excitation, while off-diagonal kernels
$(m\neq n)$ capture cross-excitation.

\paragraph{Simulation via thinning.}
We simulate marked spatiotemporal events via Ogata-style thinning with a constant
envelope $\Lambda$~\citep{ogata1981lewis}. Candidate times are proposed with
$\Delta t\sim\mathrm{Exp}(\Lambda)$; for each proposed time, we sample a candidate
mark uniformly from $\{1,\dots,M\}$ and a candidate location from a base spatial
proposal. The candidate is accepted with probability proportional to the local
conditional intensity, normalized by the envelope and proposal correction;
otherwise it is rejected.

\section{Spatiotemporal Pattern Complexity}\label{sec:complexity-pillars}
Because we control the data-generating process (DGP), each complexity pillar is tied to one of the simulator components introduced in Section~\ref{section:back}: the background field, triggering kernel, branching matrix, or spatial domain. Each pillar is defined by a generator-aligned index computed from the latent mechanism itself. We
do not estimate complexity from realized event samples, since sample-based
measures conflate structural difficulty with finite-sample noise and estimator
variability. In HawkesNest, complexity indices serve as ground-truth coordinates
for controlled sweeps; they are not intended to be identifiable from opaque
real-world datasets.

Concretely, HawkesNest constructs a multi-axis complexity grid spanning four
pillars: space--time entanglement, background heterogeneity, cross-type
interactions, and domain topology. Each pillar specifies: (i) the structural
property being isolated, (ii) a scalar mechanism that varies this property while
holding total mass, stability, and global rate fixed, and (iii) a deterministic
index in $[0,1]$ that increases with the controlled strength. This yields
synthetic corpora with mechanism-level ground truth for downstream model
evaluation.

\begin{figure}[t]
\centering
\setlength{\tabcolsep}{2pt}

\begin{subfigure}[t]{0.48\linewidth}
  \centering
  \includegraphics[width=\linewidth]{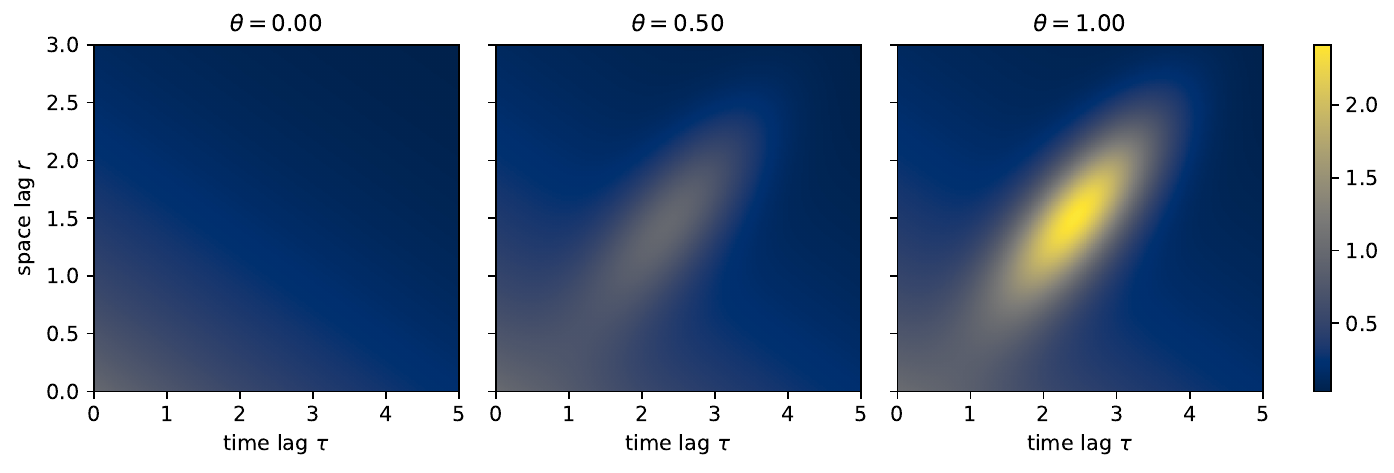}
  \caption{Increasing space--time entanglement.}
  \label{fig:pillar_examples_entanglement}
\end{subfigure}
\hfill
\begin{subfigure}[t]{0.48\linewidth}
  \centering
  \includegraphics[width=\linewidth]{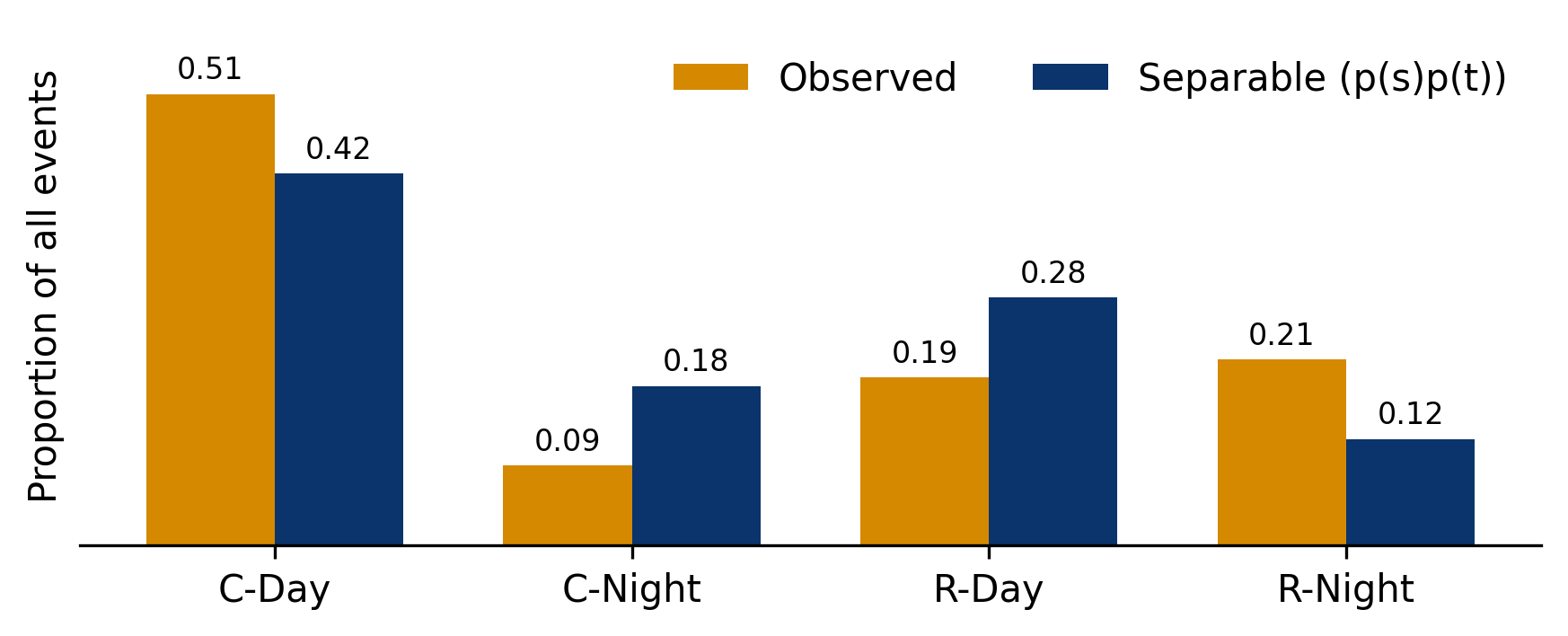}
  \caption{Separable vs. observed zone-time proportions.}
  \label{fig:pillar_examples_separable}
\end{subfigure}
\vspace{0.4em}

\begin{subfigure}[t]{0.48\linewidth}
  \centering
  \includegraphics[width=\linewidth]{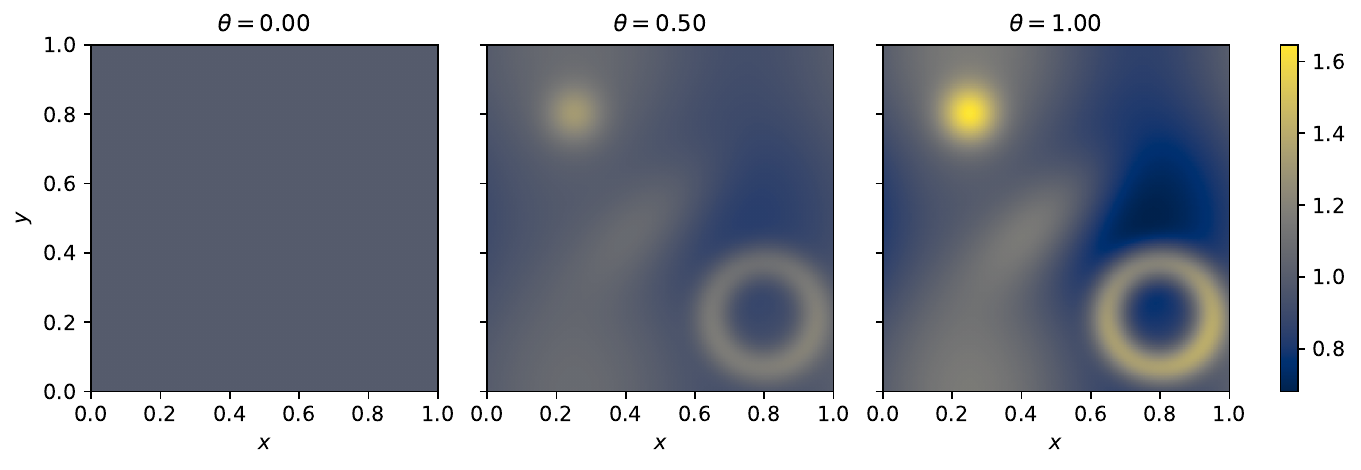}
  \caption{Increasing background heterogeneity.}
  \label{fig:pillar_examples_heterogeneity}
\end{subfigure}
\hfill
\begin{subfigure}[t]{0.48\linewidth}
  \centering
  \includegraphics[width=\linewidth]{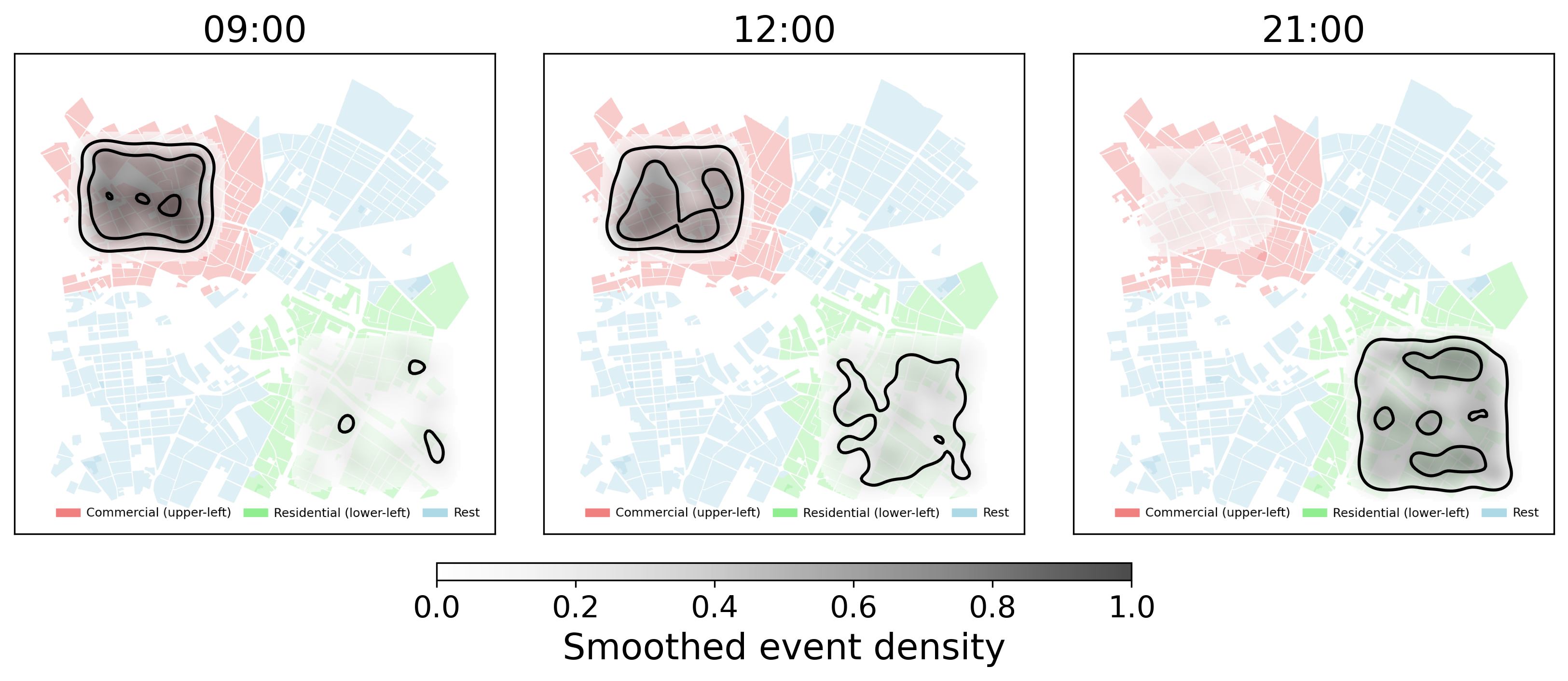}
  \caption{Drifting background-driven hot spot.}
  \label{fig:pillar_examples_kde}
\end{subfigure}

\vspace{-0.3em}

\caption{Illustrative examples of controlled spatiotemporal structure in HawkesNest.}
\label{fig:pillar_examples}

\vspace{-0.5em}
\end{figure}
\subsection{Pillar I: Space--Time Entanglement}
\label{sec:pillar-entanglement}
This pillar targets the triggering kernel introduced in Section~\ref{section:back}. Most STPP models assume that space and time interact through a separable kernel,
\[
\phi(\mathbf{s},t)=\phi_s(\mathbf{s})\,\phi_t(t),
\]
so temporal decay is identical across space and spatial dispersion is independent of event age. This is often violated when event timing affects offspring location, and vice versa. We call this departure from separability space--time entanglement.

\paragraph{Generative mechanism.}
Starting from a separable baseline $\phi_{\mathrm b}(\mathbf r,\tau)=g(\tau)h(\mathbf r)$, we introduce entanglement through a standardized non-separable modulation field $z(\mathbf r,\tau)$:
\[
\phi_{\theta}(\mathbf r,\tau)
=
c_{\theta}\,\phi_{\mathrm b}(\mathbf r,\tau)
\exp\!\bigl(\theta z(\mathbf r,\tau)\bigr),
\]
where $\theta\in[0,1]$ controls deformation strength and $c_\theta$ preserves total kernel mass. Thus temporal decay and branching ratio remain fixed while the joint space--time shape changes. At $\theta=0$, the kernel is separable; increasing $\theta$ progressively warps it along the non-separable direction defined by $z$. Figure~\ref{fig:pillar_examples_entanglement} illustrates this transition.

\paragraph{Complexity index.}
To obtain a deterministic, scale-free dependence measure, we associate each $\theta$ with a reference bivariate Gaussian $(T,X)$ with correlation $\rho(\theta)$, where $\rho(0)=0$ and $\rho$ increases monotonically. Its mutual information is
\[
I_{\mathrm G}(\theta)
=
-\tfrac12\log\!\bigl(1-\rho(\theta)^2\bigr).
\]
Normalizing over the experimental range gives
\begin{align}
\alpha_{\mathrm{ent}}
=
\frac{I_{\mathrm G}(\theta)}
     {I_{\mathrm G}(\theta_m)}
\in[0,1],
\end{align}
where $\theta\in[0,\theta_m]$ and $\theta_m<1$. Hence $\alpha_{\mathrm{ent}}=0$ only for the separable case and increases with space--time coupling, as verified in Figure~\ref{fig:exp01_mono}.

\paragraph{Example.}
Consider dispatch calls in a city with a commercial zone (C) and residential zone (R). Marginally, 60\% of calls occur in C and 40\% in R; 70\% occur during the day and 30\% at night. A separable model would combine these marginals independently. But if C dominates daytime calls and R dominates nighttime calls, the joint pattern is non-separable. Figure~\ref{fig:pillar_examples_separable} shows how the separable baseline misallocates mass across zone-time pairs.

\subsection{Pillar II: Background Heterogeneity}
\label{headings}

This pillar targets the background intensity component introduced in Section~\ref{section:back}. It captures exogenous spatial or spatiotemporal variation before self-excitation is considered. In a homogeneous baseline, the expected arrival rate is constant over the domain. In realistic event systems, population density, land use, infrastructure, mobility, and daily activity cycles create persistent spatial and temporal contrasts. HawkesNest isolates this source by varying the background intensity while keeping the triggering kernel and branching structure fixed.

\paragraph{Generative mechanism.}
Let $g(\mathbf{s},t)$ be a structured field on $\mathcal S\times[0,T]$, normalized to unit mean. We define
\begin{align}
\mu_\theta(\mathbf{s},t)
=
\lambda_0\left[(1-\theta)+\theta \,g(\mathbf{s},t)\right],
\end{align}
where $\lambda_0$ is fixed. This preserves the expected background intensity, $\mathbb E[\mu_\theta]=\lambda_0$, so heterogeneity changes only the spatial or spatiotemporal allocation of mass, not the global event rate. The triggering kernel and branching strength remain fixed. At $\theta=0$, the background is uniform; at $\theta=1$, it fully inherits the structure of $g$. Figure~\ref{fig:pillar_examples_heterogeneity} shows this ladder.

\paragraph{Complexity index \(\alpha_{\mathrm{het}}\).}
The fixed-mean constraint separates heterogeneity from event volume: increasing \(\alpha_{\mathrm{het}}\) redistributes background mass over space--time without changing the expected total background rate. We therefore measure heterogeneity by the variance of the normalized background intensity:
\begin{align}
\alpha_{\mathrm{het}}
&=
\mathbb{V}\mathrm{ar}_{\mathbf{s},t}
\left[
\frac{\mu_\theta(\mathbf{s},t)}{\lambda_0}
\right]
=
\theta^2 \,
\mathbb{V}\mathrm{ar}_{\mathbf{s},t}
\left[g(\mathbf{s},t)\right]
\nonumber\\
&=
\theta^2 \,
\mathbb E_{\mathbf{s},t}
\left[
\left(g(\mathbf{s},t)-1\right)^2
\right].
\end{align}

Thus \(\alpha_{\mathrm{het}}=0\) exactly for a homogeneous background, and it increases quadratically as background mass concentrates away from uniformity. Up to normalization, this coincides with a \(\chi^2\)-divergence between the normalized background field and the uniform reference measure~\citep{csiszar1967information}.

\paragraph{Example.}
In the dispatch example, the commercial zone can dominate during working hours while the residential zone dominates overnight. Figure~\ref{fig:pillar_examples_kde} illustrates this as a drifting background-driven hot spot from \textbf{C} to \textbf{R}.

\subsection{Pillar III: Cross-Type Interactions}\label{sec:pillar-graph}

This pillar targets the multi-mark branching matrix \(\mathbf{A}\) introduced in Section~\ref{section:back}. Cross-type interactions capture endogenous coupling among event types through off-diagonal excitation entries. If $\mathbf A$ is diagonal, types evolve independently. Off-diagonal entries create directed excitation pathways, producing coordinated activity, cascades, or modular triggering. HawkesNest isolates this source by varying only cross-type excitation.

\paragraph{Generative mechanism.}
We use branching matrices
\[
\mathbf A_\theta=\delta\mathbf I+\theta\mathbf B,
\]
where $\delta>0$ is a shared self-excitation term and $\mathbf B$ is a fixed block-modular off-diagonal template. Event types are partitioned into $K$ blocks with labels $z_m\in\{1,\dots,K\}$, and
\[
\mathbf B[m,n]
=
\begin{cases}
w_{\mathrm{in}}, & z_m=z_n,\\
w_{\mathrm{out}}, & z_m\neq z_n,
\end{cases}
\qquad
w_{\mathrm{in}}>w_{\mathrm{out}}\geq0,
\]
with $\mathbf B[m,m]=0$. A global rescaling enforces $\|\mathbf A\|_2\leq\eta_{\max}<1$, ensuring Hawkes stability. Thus $\theta$ controls cross-type interaction strength while preserving the interaction pattern.

\paragraph{Complexity index.}
The interaction index increases when cross-type excitation is both strong and structurally organized. Let $\|\mathbf A\|_2$ be the spectral norm and $Q(\mathbf A)$ the Newman modularity of the symmetrized interaction graph. Define
\begin{align}
\alpha_{\mathrm{int}}
=
\frac{\|\mathbf A\|_2}{\eta_{\max}}
\sqrt{
\frac{Q(\mathbf A)}{Q_{\max}(\mathbf A)}
},
\end{align}
where $Q_{\max}(\mathbf A)=1-\sum_m p_m^2$ is the modularity bound induced by node-strength proportions $p_m=s_m/\sum_j s_j$. Under $\mathbf A_\theta=\delta\mathbf I+\theta\mathbf B$, the modularity ratio is constant for $\theta>0$, while the spectral norm of the cross-type component increases monotonically. Hence $\alpha_{\mathrm{int}}$ increases from no cross-type structure to the stability-limited maximum.

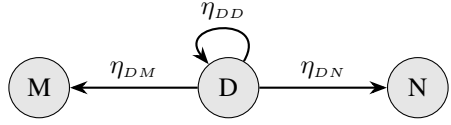
\begin{figure}[ht]
\centering
\begin{tikzpicture}[
  marknode/.style={circle, draw=black, fill=gray!20, minimum size=8mm, inner sep=0pt},
  >=Stealth,
  scenariolabel/.style={font=\bfseries}
]
\node[marknode] (D) at (0,1) {D};
\node[marknode] (M) at (-2.5,1) {M};
\node[marknode] (N) at (2.5,1) {N};

\draw[->, line width=0.8pt] (D) -- node[midway, above, sloped] {$\eta_{\scriptscriptstyle DM}$} (M);
\draw[->, line width=0.8pt] (D) -- node[midway, above, sloped] {$\eta_{\scriptscriptstyle DN}$} (N);
\draw[->, line width=0.8pt, looseness=4, min distance=4mm]
  (D) to[out=60, in=130] node[above] {$\eta_{\scriptscriptstyle DD}$} (D);
\end{tikzpicture}
\caption{Directed triggering graph for disturbances (\textbf{D}), medical calls (\textbf{M}), and noise complaints (\textbf{N}).}
\label{fig:mark-scenario}
\end{figure}

\paragraph{Example.}
Consider medical (M), disturbance (D), and noise (N) calls. Disturbances can trigger all types, while M and N do not trigger others. Stronger $D\!\to\!M$ excitation represents situations where disturbances produce medical follow-ups. Figure~\ref{fig:mark-scenario} shows the induced directed triggering graph.

\subsection{Pillar IV: Domain Topology}\label{sec:pillar-topology}

\begin{figure}[t]
  \centering
  \includegraphics[width=0.8\linewidth]{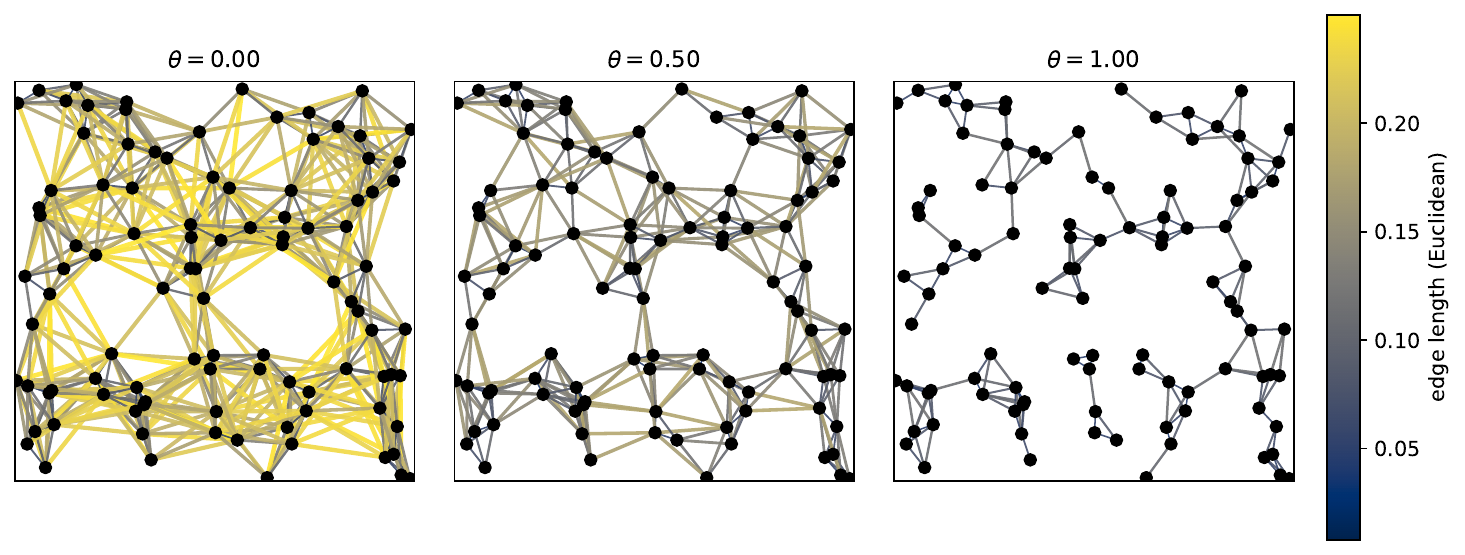}
  \caption{Random geometric graphs for increasing values of the topology parameter, illustrating progressively sparser connectivity and longer geodesic paths.}
  \label{fig:topo_ill}
\end{figure}

This pillar targets the spatial domain introduced in Section~\ref{section:back}. The previous pillars assume that event locations are embedded in a flat, fully accessible Euclidean region. Many systems violate this assumption: traffic, river, and fault-line events propagate along constrained substrates where feasible paths and geodesic distances differ from straight-line distances. HawkesNest treats topology as a property of spatial accessibility.

\paragraph{Generative mechanism.}
We represent the domain by a random geometric graph on the unit square. For topology level $\theta\in[0,1]$, we sample $n$ nodes uniformly in $[0,1]^2$ and connect nodes whose Euclidean distance is below
\[
r(\theta)=r_{\max}-\theta(r_{\max}-r_{\min}),
\qquad
0<r_{\min}<r_{\max}.
\]
We retain the largest connected component. At $\theta=0$, the graph is dense and geodesic distances are close to Euclidean distances. As $\theta$ increases, the graph becomes sparser and shortest paths become longer and less direct. Figure~\ref{fig:topo_ill} shows the resulting topology ladder.
\paragraph{Complexity index.}
Let $\mathcal G=(\mathbb V,\mathbb E)$ be the domain graph, with node positions
$\mathbf x_u\in\mathbb R^2$, and let
$\mathcal P=\{(u,v)\in\mathbb V\times\mathbb V:\ u\neq v,\ d_g(u,v)<\infty\}$
denote the connected node pairs. Define the average Euclidean and
graph-geodesic distances by
\[
\bar d_e
=
\frac{1}{|\mathcal P|}
\sum_{(u,v)\in\mathcal P} d_e(u,v),
\qquad
\bar d_g
=
\frac{1}{|\mathcal P|}
\sum_{(u,v)\in\mathcal P} d_g(u,v).
\]
The topology index is then
$\alpha_{\mathrm{top}} = 1 - \bar d_e / \bar d_g$.
Thus $\alpha_{\mathrm{top}}=0$ when graph distances match straight-line
distances, and increases as geodesic paths become more constrained.

\subsection{Discussion: Scope and Interpretation of the Pillars}

The four pillars are generator-aligned coordinates tied to specific latent
components of the DGP. They operationalize entanglement, heterogeneity, cross-type interaction, and topology as controlled mechanism families within the HawkesNest simulator.

The indices are meaningful within a fixed generative family. Within such a family, the control parameter defines a quantitative sweep coordinate. Changing the background field family, kernel family, interaction graph family, or domain family changes the experimental regime itself; such changes are qualitative design choices rather than points on a single universal complexity scale. HawkesNest can instantiate these regimes, but its indices are intended for controlled comparisons within a specified mechanism family.

This separation is deliberate: HawkesNest supports controlled sweeps within a
fixed structural family while allowing users to substitute alternative
backgrounds, kernels, interaction graphs, or domains. It should therefore be
viewed as a diagnostic tool for stress-testing STPP models under known
structural difficulty, and as a modular platform for exploring alternative
generative assumptions.
\section{Experiments and Results}

\paragraph{Experimental protocol and data regimes.}
We evaluate whether controlled structural complexity exposes model-specific
failure modes.
All datasets are generated with fixed global rate, branching stability,
simulation budget, and train/test split policy.
We use two stress-test regimes.
The joint regime $(\mathrm{C}_1)$--$(\mathrm{C}_5)$ varies background
heterogeneity and space--time entanglement along the diagonal
\begin{equation*}
  (\theta_{\mathrm{het}},\,\theta_{\mathrm{ent}})
  \;\in\;
  \bigl\{
    (0,0),\;
    (0.25,\,0.25),\;
    (0.5,\,0.5),\;
    (0.75,\,0.75),\;
    (1,1)
  \bigr\},
\end{equation*}
with $\theta_{\mathrm{top}}=0.1$ and $\theta_{\mathrm{int}}=0$.
The isolated-entanglement regime $(\mathrm{L}_0)$--$(\mathrm{L}_3)$ varies
the triggering-kernel entanglement level while keeping the remaining pillars
fixed.
We report mean $\pm$ standard deviation over independent seeds.
Index-validation results are reported in Appendix~\ref{app:index-validation},
and compute details are provided in Appendix~\ref{app:compute}.
The simulator, configuration files, notebooks, and reproduction scripts are
provided in the public code release at \anonrepo{}
(code version \texttt{v0.1.0-arxiv}).

\paragraph{Models and baselines.}
We evaluate two model classes.
Hawkes and Cox--Hawkes are likelihood-based Hawkes-family baselines with direct
structural alignment to the generator.
AutoSTPP is a neural STPP model and is used to test whether a more flexible
learned model remains sensitive to controlled space--time entanglement.

\paragraph{Index sanity check.}
The proposed indices behave as intended: each increases monotonically when its
own control parameter is varied in isolation, and the full factorial sweep
yields near-orthogonal axes, with main effects explaining at least $99\%$ of
the variance for every index. These checks validate the simulator coordinates;
the corresponding monotonicity curves and variance decomposition are reported in
Appendix~\ref{app:index-validation}.

\begin{figure}[t]
  \centering
  \includegraphics[width=\linewidth]{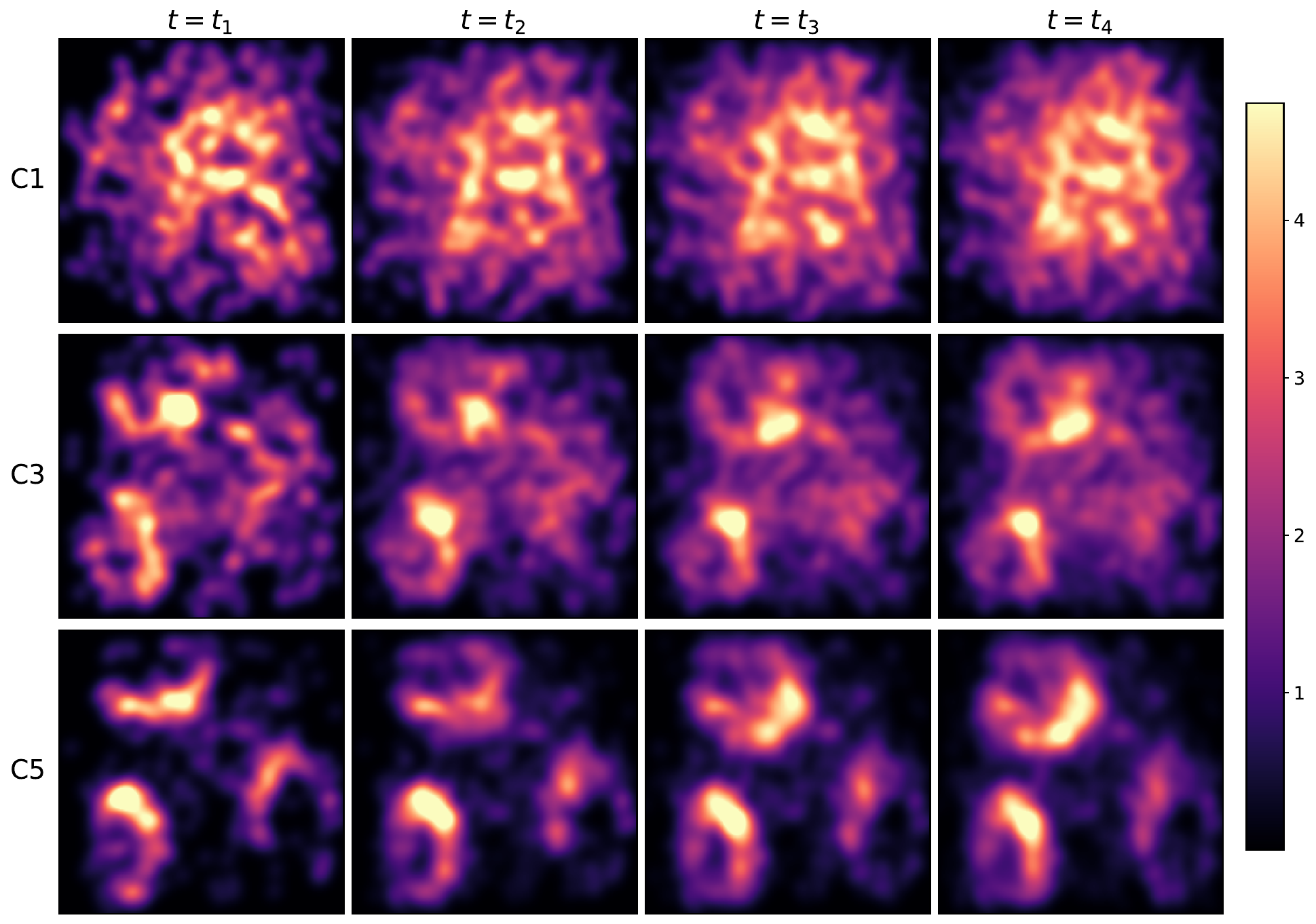}
  \caption{Kernel density estimates of realized event locations over four temporal windows for representative configurations $\mathrm{C}_1$, $\mathrm{C}_3$, and $\mathrm{C}_5$. As complexity increases, event-density patterns become less localized, more diffuse, and less stable over time.}
  \label{fig:exp02_kde}
\end{figure}

\paragraph{Hawkes-family baselines under joint complexity.}
In the \(\mathrm{C}_1\)--\(\mathrm{C}_5\) regime, we first test the model family with the strongest inductive alignment to the generator: likelihood-based Hawkes models. We evaluate Hawkes and Cox--Hawkes across the five increasing-complexity configurations defined above.
This sweep moves along the joint heterogeneity--entanglement diagonal of the design grid.
Figure~\ref{fig:exp02_kde} visualizes representative data regimes generated by this sweep, showing increasingly diffuse and less stable event-density patterns as complexity increases. Table~\ref{tab:LL} shows monotone degradation for both baselines. This is the main diagnostic result: even models structurally matched to the Hawkes backbone fail progressively as controlled complexity increases.

\begin{table}[t]
\centering
\caption{Per-event test log-likelihood for Hawkes-family baselines across joint heterogeneity--entanglement configurations $\mathrm{C}_1$--$\mathrm{C}_5$. Reported values are mean $\pm$ standard deviation over 5 runs. Higher is better.}
\label{tab:LL}
\small
\setlength{\tabcolsep}{5pt}
\begin{tabular}{lccccc}
\toprule
Model & $\mathrm{C}_1$ & $\mathrm{C}_2$ & $\mathrm{C}_3$ & $\mathrm{C}_4$ & $\mathrm{C}_5$ \\
\midrule
Cox--Hawkes &
$2.24 \pm 0.10$ &
$1.97 \pm 0.14$ &
$1.59 \pm 0.09$ &
$1.26 \pm 0.22$ &
$0.75 \pm 0.29$ \\
Hawkes &
$2.59 \pm 0.06$ &
$2.52 \pm 0.05$ &
$2.37 \pm 0.04$ &
$2.27 \pm 0.13$ &
$2.09 \pm 0.11$ \\
\bottomrule
\end{tabular}
\end{table}

\begin{table}[t]
\centering
\caption{AutoSTPP final-budget metrics under increasing entanglement levels $\mathrm{L}_0$--$\mathrm{L}_3$. Lower test NLL and higher intensity correlation are better.}
\label{tab:autostpp_nll_corr}
\small
\setlength{\tabcolsep}{4.5pt}
\begin{tabular}{lcccc}
\toprule
Metric & $\mathrm{L}_0$ & $\mathrm{L}_1$ & $\mathrm{L}_2$ & $\mathrm{L}_3$ \\
\midrule
Test NLL &
$-1.832 \pm 0.001$ &
$-1.824 \pm 0.005$ &
$-1.826 \pm 0.001$ &
$-1.779 \pm 0.004$ \\
Intensity corr. &
$0.732 \pm 0.010$ &
$0.687 \pm 0.031$ &
$0.672 \pm 0.012$ &
$0.580 \pm 0.024$ \\
\bottomrule
\end{tabular}
\end{table}

\paragraph{AutoSTPP under isolated entanglement.}
We next test whether the same diagnostic principle exposes failures in a neural STPP model. We evaluate AutoSTPP across isolated entanglement levels $\mathrm{L}_0$--$\mathrm{L}_3$, increasing $\theta_{\mathrm{ent}}$ while holding the other pillars fixed. The results reveal a layered failure mode. Table~\ref{tab:autostpp_nll_corr} shows that final test NLL only weakly reflects the increasing difficulty: performance is substantially worse at $\mathrm{L}_3$ than at $\mathrm{L}_0$, but the intermediate levels are not strictly monotone. In contrast, the intensity-correlation diagnostic reveals a clear structural degradation, dropping steadily from $0.732$ at $\mathrm{L}_0$ to $0.580$ at $\mathrm{L}_3$. This indicates that likelihood alone can partially hide failures in recovering the latent generative structure, whereas correlation with the ground-truth intensity exposes the effect of increasing space--time entanglement more directly.

\begin{figure}[t]
  \centering
  \includegraphics[width=0.6\linewidth]{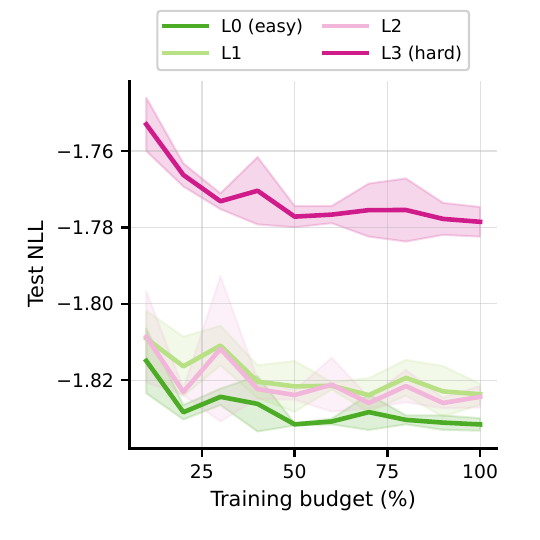}
  \caption{AutoSTPP under increasing entanglement levels $\mathrm{L}_0$--$\mathrm{L}_3$. Lines show mean test NLL and bands show variability over seeds. Lower is better.}
  \label{fig:autostpp_budget}
\end{figure}

Figure~\ref{fig:autostpp_budget} adds a third diagnostic signal. As the training budget increases from $10\%$ to $100\%$, the gap between low-entanglement regimes narrows in NLL terms, but the high-entanglement regime remains separated throughout, suggesting that additional training budget does not remove the difficulty. In particular, $\mathrm{L}_3$ does not catch up even at full budget, consistent with an inductive-bias or optimization limitation under strong entanglement. Taken together, these results show that HawkesNest exposes two distinct neural failure modes under entanglement: degraded statistical fit and degraded recovery of the latent generative structure.

\section{Discussion and Outlook}

HawkesNest is designed for diagnostic stress testing, not leaderboard-style
benchmarking.
The index-validation experiments establish the necessary construct validity:
each index increases under its own control parameter, and the variance
decomposition shows that cross-pillar interactions are negligible over the
tested grid.
This matters because model failures can be interpreted against known latent
mechanisms, rather than against uncontrolled shifts in sample size, global
rate, or branching stability.

The empirical results illustrate this diagnostic role at two levels.
The joint heterogeneity--entanglement sweep $\mathrm{C}_1$--$\mathrm{C}_5$
induces monotone degradation in Hawkes and Cox--Hawkes --- a conservative
stress test, since these models have the closest inductive alignment to the
generator, so their degradation indicates a genuine modeling limitation rather
than a family mismatch.
AutoSTPP exposes a qualitatively different failure mode under isolated
entanglement: test NLL degrades, correlation with the ground-truth intensity
drops, and additional training budget does not close the high-entanglement gap.
HawkesNest thus reveals both statistical degradation and degraded recovery of
latent DGP structure.

A model's performance profile across axes can indicate which structural
assumption is failing: separability of triggering, homogeneity of background
intensity, independence of event types, or Euclidean accessibility of the
domain.
The framework complements real-world benchmarks: opaque datasets remain
necessary for external validity, but controlled synthetic regimes are better
suited for attributing failure to specific mechanisms.

\paragraph{Limitations.}
The proposed indices are generator-aligned coordinates, not universal measures
of real-world complexity.
They are meaningful within a fixed mechanism family and should not be
interpreted as scalar rankings across arbitrary DGPs.
The current experiments evaluate a limited set of model families and focus on
likelihood and intensity-recovery diagnostics; downstream operational metrics
remain outside the present scope.
These limitations are deliberate: HawkesNest isolates mechanisms first, before
expanding to broader application-specific evaluations.

\paragraph{Future work.}
A natural next step is to enrich the generator within each pillar.
For heterogeneity, this could include background fields governed by PDE or SDE
dynamics, allowing exogenous rates to diffuse, advect, or fluctuate
stochastically over space and time.
More broadly, HawkesNest can incorporate contextual fields external to the point
process --- weather, mobility, land use, population density, or
remote-sensing covariates --- and specify how these fields modulate event
intensity, triggering, and spatial accessibility.
In the longer term, we envision HawkesNest evolving into a living benchmark with
mixed-pillar settings, domain-specific covariates, and user-contributed
generators, serving as a unit-test layer for STPP components before deployment
on proprietary or noisy real-world data.
\paragraph{Code.} We provide a public release with the simulator, configuration files, and experimental scripts at \anonrepo.

\begin{ack}
The authors received support from the Bundesministerium f\"ur Bildung und Forschung (BMBF) under Grant No.~01W23005 for the project EVENTFUL.
\end{ack}

\bibliographystyle{plainnat}
\bibliography{ref_v}

%
\clearpage
\appendix

\lstdefinestyle{hawkescmd}{
  basicstyle=\ttfamily\small,
  breaklines=true,
  breakatwhitespace=false,
  frame=single,
  rulecolor=\color{gray!40},
  backgroundcolor=\color{gray!5},
  keywordstyle=\color{blue!70!black},
  commentstyle=\color{green!50!black},
  stringstyle=\color{orange!80!black},
  showstringspaces=false,
  aboveskip=6pt,
  belowskip=6pt,
}
\lstset{style=hawkescmd}

\section{Software and Reproducibility Details}
\label{app:software}

HawkesNest is a configurable synthetic spatio-temporal point-process workbench built around a Hawkes-process data-generating process (DGP). The repository supports two complementary modes of use. In \emph{custom DGP mode}, users specify the ingredients of a synthetic process directly: spatial domain, background intensity, triggering kernel, mark or adjacency structure, rate and stability controls, and random-seed schedule. In \emph{recipe mode}, users call validated benchmark presets such as \texttt{EntanglementSuite} and \texttt{HeterogeneitySuite}. These suites are suite presets over the same configurable DGP layer.

The public repository is available at \anonrepo.
Code version: \texttt{v0.1.0-arxiv}.
Paper reproduction instructions are provided in \texttt{docs/paper\_reproduction.md}.
All commands below assume execution from the repository root.

\subsection{Software Architecture and Usage Modes}
\label{app:software-architecture}
\begin{figure}
\centering
\includegraphics[width=\linewidth]{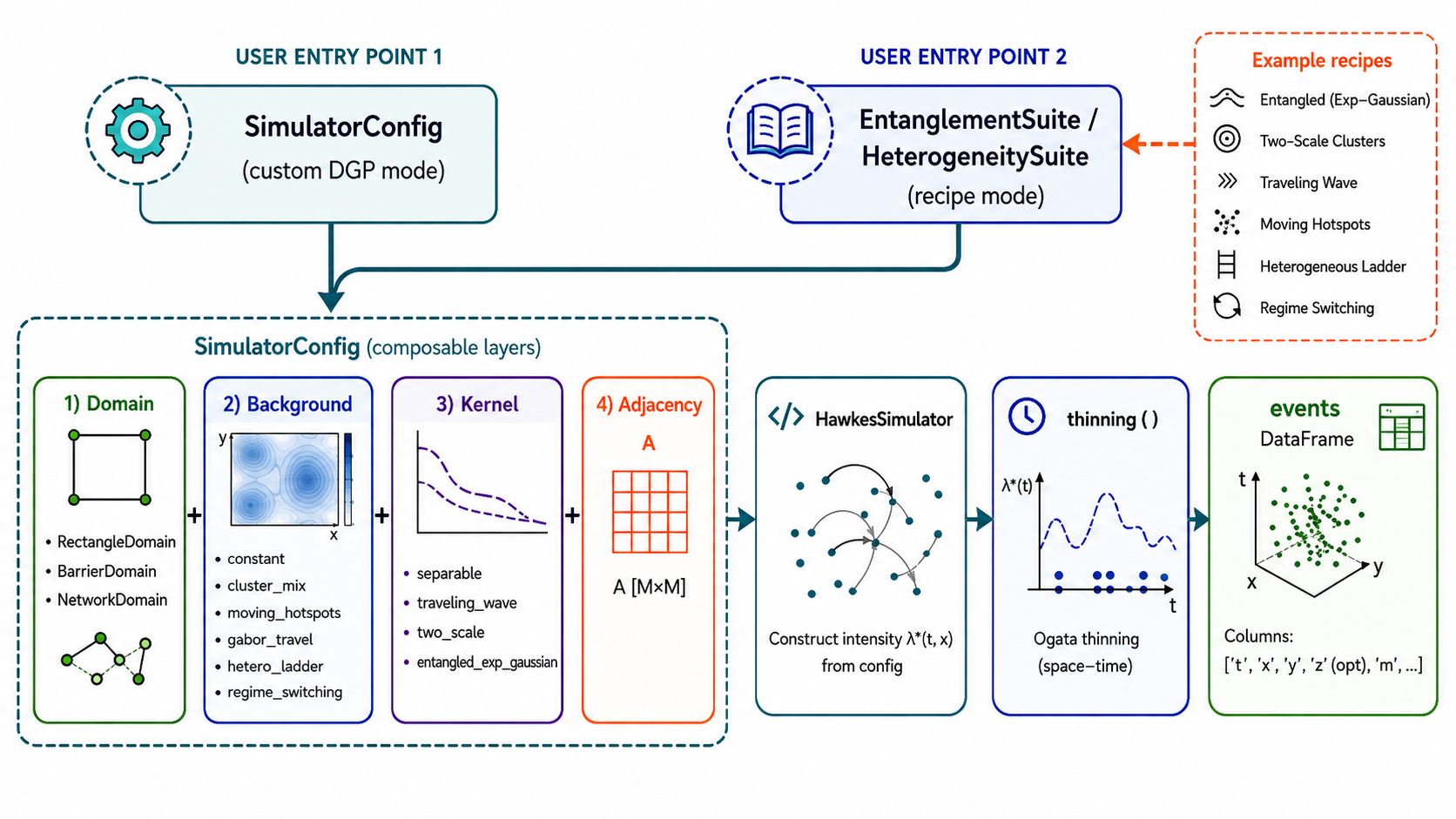}
\caption{%
  \textbf{HawkesNest software architecture.}
  Users access HawkesNest through Python APIs, command-line entry points,
  notebooks, or validated recipe interfaces.
  The DGP design layer specifies the domain, background intensity,
  triggering kernel, mark or adjacency structure, rate and stability controls,
  and seed schedule.
  These choices are resolved by \texttt{SimulatorConfig} into a
  \texttt{HawkesSimulator}, which generates event streams by thinning.
  Outputs include event tables, JSONL sequences, metadata, intensity grids,
  complexity-index summaries, and diagnostic figures.
  Visualization and validation utilities support event-cloud inspection,
  intensity snapshots, KDE/GIF diagnostics, complexity-index analysis,
  and downstream likelihood evaluation.
}
\label{fig:hawkesnest_architecture}
\end{figure}
Figure~\ref{fig:hawkesnest_architecture} summarizes the HawkesNest workflow.
The architecture separates six roles:
\begin{enumerate}
  \item \textbf{DGP design layer.} Defines the synthetic process (domain, background, kernel, adjacency).
  \item \textbf{Configuration layer.} Resolves the design into executable simulator objects via
        \texttt{SimulatorConfig}, config factories, and recipe presets.
  \item \textbf{Simulator layer.} Combines exogenous background intensity and endogenous
        triggering intensity, then generates events by thinning.
  \item \textbf{Recipe layer.} Exposes validated benchmark presets.
  \item \textbf{Artifact layer.} Writes event tables, sequence files, metadata, intensity grids,
        complexity-index summaries, and figures.
  \item \textbf{Visualization and validation layer.} Supports qualitative inspection
        and quantitative checks.
\end{enumerate}

The benchmark suites are \texttt{EntanglementSuite} and
\texttt{HeterogeneitySuite}.
The former exposes the isolated space--time entanglement ladder
$\mathrm{L}_0$--$\mathrm{L}_3$;
the latter exposes the background-heterogeneity ladder
$\mathrm{H}_0$--$\mathrm{H}_3$.
Both recipes invoke the same simulator stack:
\[
  \texttt{SimulatorConfig}
  \;\longrightarrow\;
  \texttt{HawkesSimulator}
  \;\longrightarrow\;
  \texttt{thinning}.
\]
The joint $\mathrm{C}_1$--$\mathrm{C}_5$ regimes used in the paper are not
separate recipe classes.
They are paper-defined experimental regimes obtained by varying complexity
coordinates within the same configurable simulation framework.

\subsection{Generating a Custom Dataset}
\label{app:single-custom}

A custom synthetic process can be generated directly via \texttt{SimulatorConfig}.
The example below uses a rectangular domain, a clustered background intensity,
a separable triggering kernel, and a one-mark excitation matrix.

\begin{lstlisting}[language=Python]
from hawkesnest.config import SimulatorConfig

config = {
    "domain": {
        "type": "rectangle",
        "x_min": 0.0, "x_max": 1.0,
        "y_min": 0.0, "y_max": 1.0,
    },
    "backgrounds": [{
        "type": "function",
        "name": "cluster_mix",
        "centers": [[0.25, 0.25], [0.75, 0.65]],
        "sigma": 0.12,
        "a0": 0.2,
        "amp": 1.5,
    }],
    "kernels": [{
        "type": "separable",
        "temporal_decay": 0.4,
        "spatial_sigma": 0.12,
    }],
    "adjacency": [[0.20]],
    "lambda_max": 25.0,
}

cfg = SimulatorConfig.model_validate(config)
events, parents = cfg.build().simulate(
    n=100,
    seed=7,
    tau_max=5.0,
    debug=False,
)
\end{lstlisting}

The main controls are the spatial domain, background family, triggering kernel,
adjacency matrix, thinning envelope $\lambda_{\max}$,
event budget $n$, time horizon $\tau_{\max}$, and random seed.
Marked processes are obtained by providing multiple background components and a
matrix-valued adjacency structure,
enabling self-excitation, asymmetric cross-excitation,
or nearly independent mark channels without changing the simulator backend.

\subsection{Generating Validated Benchmark Recipes}
\label{app:validated-recipes}

The benchmark suites provide a short path to reproducible benchmark data.

\medskip
\noindent\textbf{Python API.}\par\vspace{2pt}
\begin{lstlisting}[language=Python]
from hawkesnest.suites import EntanglementSuite, HeterogeneitySuite

ent = EntanglementSuite().generate(level="L2", n_events=50, seed=123)
het = HeterogeneitySuite().generate(level="H3", n_events=50, seed=123)
\end{lstlisting}

\medskip
\noindent\textbf{Command-line API.}\par\vspace{2pt}
\begin{lstlisting}[language=bash]
python -m hawkesnest.cli generate entanglement \
    --level L2 --n-events 50 --seed 123 \
    --out outputs/entanglement_demo

python -m hawkesnest.cli generate heterogeneity \
    --level H3 --n-events 50 --seed 123 \
    --out outputs/heterogeneity_demo
\end{lstlisting}

A small multi-level corpus can be generated with:
\begin{lstlisting}[language=bash]
python -m hawkesnest.cli generate-corpus entanglement \
    --levels L0 L1 L2 L3 \
    --seeds 0 1 \
    --n-events 50 \
    --out outputs/entanglement_corpus
\end{lstlisting}

The entanglement recipe uses a separable triggering kernel at $\mathrm{L}_0$
and increasingly nonseparable traveling-wave kernels for
$\mathrm{L}_1$--$\mathrm{L}_3$.
The heterogeneity recipe varies the background field across
$\mathrm{H}_0$--$\mathrm{H}_3$ while keeping the Hawkes simulator stack fixed.

\subsection{Paper Regimes and Complexity Sweeps}
\label{app:paper-suites}

The paper uses three synthetic experimental regimes:
the joint heterogeneity--entanglement stress test $\mathrm{C}_1$--$\mathrm{C}_5$,
the isolated entanglement sweep $\mathrm{L}_0$--$\mathrm{L}_3$,
and the four-dimensional index-validation sweep.

\medskip
\noindent\textbf{Joint heterogeneity--entanglement regimes
$(\mathrm{C}_1)$--$(\mathrm{C}_5)$.}\quad
The main predictive-stress experiment evaluates how per-event test log-likelihood
changes as structural complexity increases while global simulation controls are held fixed.
The five configurations sweep
\begin{equation*}
  \bigl(\theta_{\mathrm{het}},\,\theta_{\mathrm{ent}}\bigr)
  \;\in\;
  \bigl\{
    (0,\,0),\;
    (0.25,\,0.25),\;
    (0.5,\,0.5),\;
    (0.75,\,0.75),\;
    (1,\,1)
  \bigr\},
\end{equation*}
with topology and interaction parameters fixed at
$\theta_{\mathrm{top}} = 0.1$ and $\theta_{\mathrm{int}} = 0$.
The background intensity is time-varying, and the triggering kernel uses a
multiplicative $(1 + \mathbf{r}\tau)$ coupling between spatial displacement
$\mathbf{r}$ and event age $\tau$, creating joint space--time structure in
both the exogenous background and the endogenous triggering process.

For each $\mathrm{C}_k$, event sequences are generated under the same rate,
stability, and splitting policy.
Hawkes and Cox--Hawkes baselines are then trained and evaluated under identical
conditions using per-event test log-likelihood.
The qualitative KDE panels show that higher-complexity configurations become
less spatially localized and less temporally stable,
matching the systematic degradation in likelihood reported in the main text.

\medskip
\noindent\textbf{Isolated entanglement sweep.}\quad
The isolated entanglement experiment varies space--time entanglement while
holding the remaining structural controls fixed.
The public recipe command is:
\begin{lstlisting}[language=bash]
python -m hawkesnest.cli generate-corpus entanglement \
    --levels L0 L1 L2 L3 \
    --seeds 0 1 2 3 4 \
    --n-events <N_EVENTS> \
    --out outputs/entanglement_suite
\end{lstlisting}

\medskip
\noindent\textbf{Four-dimensional index-validation sweep.}\quad
The index-validation experiment varies all four structural controls over a
balanced factorial grid.
The full run contains $3125 = 5^4 \times 5$ rows
($5^4$ grid cells with five replicates per cell).
The grid is
\begin{align*}
  \theta_{\mathrm{het}} &\in \{0,\; 0.25,\; 0.5,\; 0.75,\; 1.0\}, \\
  \theta_{\mathrm{ent}} &\in \{0,\; 0.25,\; 0.5,\; 0.75,\; 0.95\}, \\
  \theta_{\mathrm{top}} &\in \{0,\; 0.25,\; 0.5,\; 0.75,\; 1.0\}, \\
  \theta_{\mathrm{int}} &\in \{0,\; 0.25,\; 0.5,\; 0.75,\; 0.95\}.
\end{align*}
The generation entry point is \texttt{scripts/sweeps\_complexity.py};
in full-sweep mode it dispatches to \texttt{run\_complexity\_sweep\_4d}
in \texttt{hawkesnest/experiments/complexity.py}.
The original run used $T=50$, $\lambda_0=40$, five replicates,
$\tau_{\max}=0.5$, \texttt{field\_name\_bg = moving\_gauss\_slow},
\texttt{ent\_option = rt}, and \texttt{seed\_offset = 1234}.
Index computation and variance decomposition results are reported in
Section~\ref{app:index-validation}.

\subsection{Index Validation}
\label{app:index-validation}

\begin{figure}[h]
  \centering
  \includegraphics[width=0.75\linewidth]{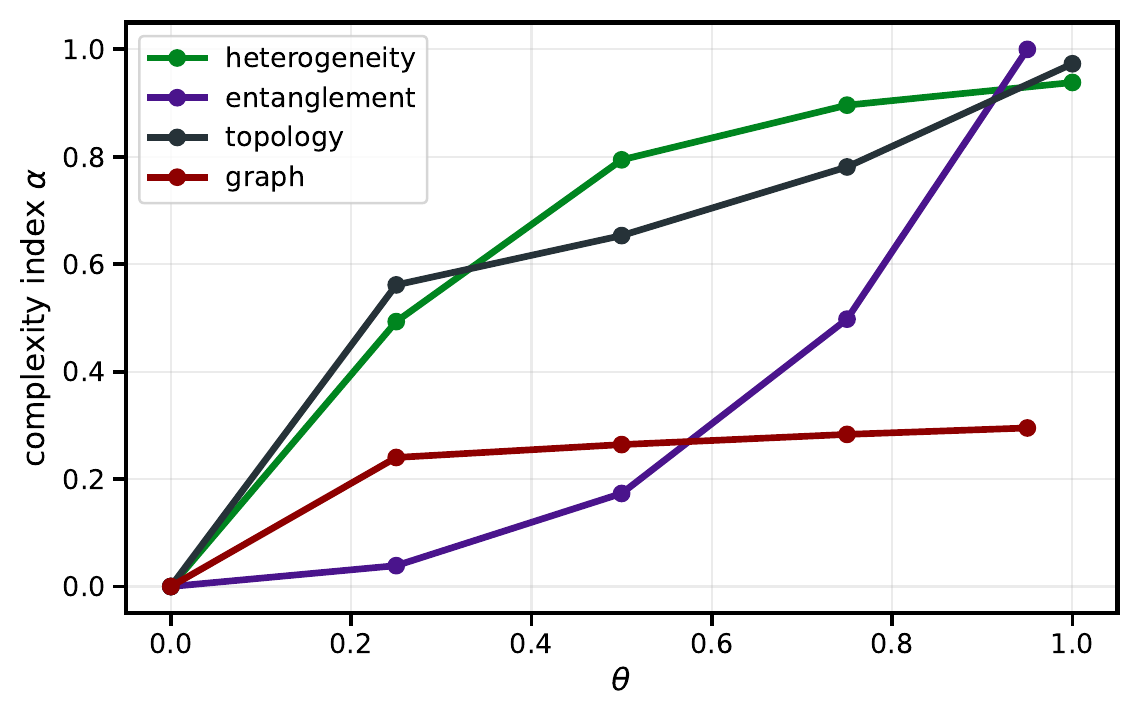}
  \caption{%
    \textbf{Monotonicity of complexity indices.}
    Each index $\alpha$ increases monotonically with its corresponding control
    parameter $\theta$ when varied in isolation, with all other parameters held
    at baseline. The curves confirm that each complexity knob has a clean,
    near-independent effect on its target index.
  }
  \label{fig:exp01_mono}
\end{figure}

We validate the complexity coordinates by varying each control parameter in
isolation and by decomposing variance over the full factorial sweep.
The monotonicity curves in Figure~\ref{fig:exp01_mono} confirm that each index
responds to its intended knob while remaining insensitive to the others.
Table~\ref{tab:anova_decomp} shows that main effects explain essentially all
variance, indicating near-orthogonal control axes over the tested grid.

\begin{table}[h]
\centering
\caption{%
  \textbf{Variance decomposition over the 4D factorial sweep.}
  Fractions of $SS_{\mathrm{total}}$ attributable to the main effect of each
  index's intended knob versus all higher-order cross-pillar interactions
  ($\geq$2-way).
  Main effects dominate; cross-pillar interactions are negligible,
  supporting the near-orthogonality of the four complexity coordinates.
}
\label{tab:anova_decomp}
\small
\setlength{\tabcolsep}{8pt}
\begin{tabular}{lccc}
\toprule
Index $\alpha$ & Main effect & $\geq$2-way interactions & Total interaction \\
\midrule
$\alpha_{\mathrm{het}}$ & $0.990$ & $0.010$ & $0.010$ \\
$\alpha_{\mathrm{ent}}$ & $1.000$ & ${\approx}0$ & ${\approx}0$ \\
$\alpha_{\mathrm{top}}$ & $1.000$ & ${\approx}0$ & ${\approx}0$ \\
$\alpha_{\mathrm{int}}$ & $1.000$ & ${\approx}0$ & ${\approx}0$ \\
\bottomrule
\end{tabular}
\end{table}

\subsection{Output Format and Metadata}
\label{app:output-format}

\medskip
\noindent\textbf{CSV format.}\par
One row per event:
\begin{lstlisting}
t, x, y, m, is_triggered
\end{lstlisting}
where \texttt{t} is event time, \texttt{x} and \texttt{y} are spatial coordinates,
\texttt{m} is the mark/type label,
and \texttt{is\_triggered} records whether the accepted event had nonzero
triggering intensity.

\medskip
\noindent\textbf{JSONL format.}\par
One sequence per line:
\begin{lstlisting}[language=Python]
{
    "times":      [...],
    "locations":  [[x1, y1], [x2, y2], ...],
    "marks":      [...],
    "is_triggered": [...]
}
\end{lstlisting}

\medskip
\noindent\textbf{Metadata.}\par
Metadata exports contain the suite name, level, seed,
requested and realized event counts, simulator class,
configuration dictionary, simulation controls, and export paths.
Paper-generation scripts additionally write intensity grids,
complexity indices, diagnostic figures, and tables.

\subsection{Visualization and Diagnostics}
\label{app:visualization}

Visualization is a central part of the HawkesNest workflow.
It is used to check whether the intended DGP structure is visible in the
generated event stream and whether complexity changes produce the expected
qualitative regimes.
The public notebooks show 2D event clouds, 3D $(x,y,t)$ event clouds,
intensity snapshots, and suite-level comparisons.
Script-level utilities additionally produce KDE/GIF diagnostics for
temporal evolution.

A minimal visualization command is:
\begin{lstlisting}[language=bash]
python -m hawkesnest.cli visualize \
    outputs/entanglement_demo/events.jsonl \
    --kind space-time \
    --out outputs/entanglement_demo/space_time.png
\end{lstlisting}
The main paper figures use KDE and intensity-style panels because they show
spatial localization, temporal drift, and regime instability more clearly
than raw event scatter alone.

\subsection{Compute Requirements}
\label{app:compute}

Small examples and benchmark-suite generation runs are CPU-only and run interactively; GPU acceleration is not required for data generation. Runtime scales with event budget or time horizon $\tau_{\max}$, thinning envelope $\lambda_{\max}$, number of marks, domain size, number of seeds, and number of suite levels. The full four-dimensional index-validation sweep ($3125$ runs) took approximately $62$ CPU-hours on a single core. Parallel execution reduces wall-clock time proportionally to available cores, subject to filesystem and plotting overhead.
\end{document}